\documentclass{article}
\usepackage[preprint]{neurips_2019}

\usepackage[utf8]{inputenc} % allow utf-8 input
\usepackage[T1]{fontenc}    % use 8-bit T1 fonts
\usepackage[colorlinks=true, citecolor=blue]{hyperref}
\usepackage{url}            % simple URL typesetting
\usepackage[utf8]{inputenc} % allow utf-8 input
\usepackage{booktabs}       % professional-quality tables
\usepackage{amsfonts}       % blackboard math symbols
\usepackage{nicefrac}       % compact symbols for 1/2, etc.
\usepackage{microtype}      % microtypography
\usepackage{graphicx}
\usepackage[]{algorithm2e}
\usepackage{placeins}       % for barriers for floats 
\usepackage{caption}        % for caprion settins 
\usepackage{graphics}

\usepackage[intlimits]{amsmath}
\usepackage{amsfonts, amssymb,amscd, amsthm}

\usepackage{algorithmic}

\usepackage{natbib}
\setcitestyle{authoryear,round,semiscolon}

\usepackage{multicol}
\usepackage{multirow} % http://ctan.org/pkg/multirow
\usepackage{hhline} % http://ctan.org/pkg/hhline

\title{Factorized MultiClass Boosting}

\author{%
Igor E.~Kuralenok\\
JetBrains Research\\
St. Petersburg, Russia\\
\texttt{ikuralenok@gmail.com} \\
\And
Yurii Rebryk\\
Higher School of Economics\\
St. Petersburg, Russia\\
\texttt{y.a.rebryk@gmail.com} \\
\And
Ruslan Solovev \\
Yandex \\
St. Petersburg, Russia\\
\texttt{solovevr@gmail.com} \\
\And
Anton Ermilov\\
Higher School of Economics\\
St. Petersburg, Russia\\
\texttt{anton.yermilov@gmail.com} \\
}

\begin{document}
\maketitle

\begin{abstract} 
In this paper, we introduce a new approach to multiclass classification problem. We decompose the problem into a series of regression tasks, that are solved with CART trees. The proposed method works significantly faster than state-of-the-art solutions while giving the same level of model quality. The algorithm is also robust to imbalanced datasets, allowing to reach high-quality results in significantly less time without class re-balancing.
\end{abstract} 

\section{Introduction}
\label{introduction}
Classification with more than two classes is a natural setting for many tasks including text classification, object recognition, protein folding. High demand generated many methods whose performances vary from task to task~\citep{SOTA}. In most cases, the family of gradient boosted decision trees (GBDT) works best off-the-shelf. There are three major exceptions:
\begin{itemize}
    \item datasets where feature extraction is not trivial (images, signals, etc.). Deep learning methods have better results on this data;
    \item tiny datasets, where RBM~\citep{RBM}, ELM~\citep{ELM} are better;
    \item problems with large number of classes where performance issues make usage of GBDT impractical.
\end{itemize} 
In this paper, we focus on the last case. We propose a factorization method that makes it practical to use GBDT models for problems with a large number of classes without having to do class balancing.

There are two main approaches to apply boosting to multiclass classification problems known from the literature: task decomposition and weights boosting. The most straightforward approach is to decompose the multiclass classification problem into a system of independent binary classification problems (e.g. one vs. rest). A generalization of this approach is provided in \citet{allwein00reducing}. This approach has nice theoretical properties and has its successors \citep{NIPS2018_7953}, but this setting does not fit for dynamic nature of boosting procedure. Another way of decomposition, which is more natural for boosting, is to build a series of weak classifiers correcting each other. This approach is represented in wide variety of papers \citep{Schapire1999,SAMME} and is generalized on top of game theory principles in \citet{journals/jmlr/MukherjeeS13}. Unfortunately, these methods are rarely used in practice because they are generally outperformed by off-the-shelf GBDT libraries \citep{SOTA}. 

One more way to do boosting for multiclass classification is to build a decision vector function, whose components represent the weights of each class for provided example. The most simple way to build such a vector function is to optimize multinomial logistic regression loss (MLR) \citep{GLM} or another loss that contains information on class relationships \citep{Friedman98additivelogistic}. Attractive property of this approach is that it works well even with imbalanced classes~\citep{imbalancing_classes1,imbalancing_classes2}. The boosting procedure in this case is applied to vector function components that are built either independently or share components of ensemble elements.

When the number of classes increases, a constructing vector function becomes complicated. We can reduce complexity and the computational ``weight'' of the decision function by sharing certain components. The most popular functional basis for boosting methods are decision trees. To reduce the complexity of the vector function, we can share the same tree topology between all classes and express their differences in the leaf values only. Because of this structure (single topology and vector of class weights in the leaves) they sometimes are referenced as vector-trees. This optimization is presented in both academic studies \citep{Sun:2012:AAO:3042573.3042675,pmlr-v70-si17a} and in industrial libraries \citep{CatBoost, XGBoost}. The techniques used to build a topology vary, but it is always more computationally intensive than single classifier or regression tree building because we need to take into account information on all of the classes. To the best of our knowledge, the only exception from this rule is when we choose single ``basis'' class during optimization, but this simplification gives poor quality results.

In our work we combine both approaches. At first, we decompose the multiclass classification task into a series of \textit{regression} tasks using MLR gradient matrix factorization. Then, for each regression task we build a decision tree with state-of-the-art algorithms. Finally, we add all vectors together to get boosted classes weights. This way we are able to get lightweight, less complex models that are fast to train and robust to class imbalance. The main contributions of this paper are the following:
\begin{itemize}
    \item method for fast rank-one matrix decomposition, when one of the matrix dimensions dominates the other (SALS);
    \item method based on a decomposition of the multiclass classification task to a series of regressions;
    \item extensive comparison of the method with off-the-shelf libraries and alternative approaches.
\end{itemize}

% TODO: change it
The later sections of the paper are organized as follows: in the motivation part, the idea of the method is explained on the basis of existing approaches, the algorithm is described in details in Section~\ref{fmcb}, experimental part of the research and computation performance analysis follows in Section~\ref{experiments}.

\section{Motivation}
Many algorithms in ML solve binary classification or regression tasks. The idea of employing them for multiclass classification looks attractive. The modeling matrix approach formulated by Allwein, Schapire, and Singer \citep{allwein00reducing} allows one to split this complex task into binary classification subtasks. Unfortunately, the elements of the modeling matrix must be from a set of $\{-1,0,1\}$. This limitation makes the optimization of this matrix for a particular task much more difficult \citep{modelmatrix2,Zhao13sparseoutput}. We want to get rid of this limitation and build the modeling matrix iteratively, column by column. On each step, we greedily optimize the classification quality on the whole dataset. This approach is similar to~\citep{journals/jmlr/MukherjeeS13}, but we use regression instead of classification. Thus the mathematical grounds are completely different.

Another valuable property of the modeling matrix approach is that it fits well when the number of classes is significant, and it is hard or even impossible to train scoring function for each of them. In this case, we ``code'' the class by modeling matrix row. Then it is possible to estimate the class probability, using results of the binary classifiers defined by each column. The minimal number of the classifiers needed to code a fixed number of classes $K$ is $log K$, and this theoretically allows us to use much fewer models than the number of classes. Unfortunately, coding matrix optimization is hard, and one bad classifier can ruin the quality of the whole decision function.

Our idea is to go the opposite way: instead of minimization of the number of classifiers, we use a broad set of weak models and associate with them ``coding vectors''. A large number of weak models allows reduce the variance of an entire ensemble. Reuse of the same regression trees across classes reduces computational and informational weight of the decision. The number of classes influence only ``coding vectors'', this allows to have the arbitrary number of weak models, even less than the number of the classes. This property becomes useful when the number of classes is huge ($>1000$).

\section{Algorithm}
\label{fmcb}
In this section we describe the proposed method. To introduce the notation we start from general schema of the multinomial logistic regression optimization, the core of the method is described then followed by convergence analysis, the final part contains description of SALS method used for gradient matrix factorization.

\subsection{Multinomial logistic regression boosting}
The boosting of multinomial logistic regression is a well studied topic \citep{Friedman98additivelogistic}. If one fixes a decision vector function $H$ of $K-1$ dimensions ($H_c$ is the $c$-th class component of $H$), the probability of class $c$ for feature vector $x$ will take the following form:
\begin{equation}
\label{eq:MLRP}
P(Y = c \arrowvert x, H) = \frac{e^{I\{c < K\} H_{c}(x)}}{1+\sum_{j=1}^{K-1} e^{H_{j}(x)} }
\end{equation}
where $I$ is indicator function. One can maximize the log likelihood of the training sample:
\begin{equation}
\begin{array}{rl}
    L(\mathbf{X} \arrowvert H) &= \sum_{(x,y) \in \mathbf{X}} \log P(Y = y \arrowvert x, H) \\
    % & = \sum_{(x,y) \in \mathbf{X}} \log \frac{e^{I\{y < K\} H_y(x)}}{1+\sum_{c=1}^{K-1} e^{H_c(x)}}
    \label{eq:MLR_model}
\end{array}
\end{equation}
with dataset $\mathbf{X}$ consisting of $(x, y)$ pairs, where $y$ is the correct class for the associated item $x$. The decision function is a sum of $T$ components $H(x) = \sum_{t=1}^T h_t(x)$. Construction of the decision function is done iteratively, one $h_{t+1}$ at a time.

At each step, we want the matrix $h_{t + 1}(X)$, where $X$ is a joined training data feature matrix, to be aligned with the matrix of partial derivatives for each example (row), for each class (column) at the point of current boosting cursor $H_t(X)$:
\begin{equation}
\label{eq:grad_matrix}
\nabla L = 
\left(
\begin{array}{ccc}
      \frac{ \partial L }{ \partial H_{1}(x_{1}) } 
    & \cdots 
    & \frac{ \partial L }{ \partial H_{K-1}(x_{1}) } \\

      \vdots & \ddots & \vdots\\

      \frac{ \partial L }{ \partial H_{1}(x_{|X|}) } 
    & \cdots 
    & \frac{ \partial L }{ \partial H_{K-1}(x_{|X|}) }
\end{array}
\right)_{|X| \times (K-1)}
\end{equation}
The particular value of $\nabla L$ is calculated at each step $\nabla L_t = \nabla L(H_t(X))$. To provide alignment we look for $h_{c,t+1}$ minimizing Frobenius norm of the residue matrix:
\begin{equation}
{h}_{t+1} = \arg \min_{h} \|{h}(X) - \nabla L_t\|_F
\label{eq:MLR_boosting_step}
,\end{equation}
where $h(X)$ is matrix of $h_c(x_i), i \in \left[1,..,m\right]$ components. If we optimize $h_c$ components independently, there are $K-1$ regression trees are built at this step.

This part is the most computationally intensive for any MLR implementation and its further optimization is definitive. There are wide variety of techniques used here, but the most important problems are: scoring function for tree topology and leaf vector optimization strategy. Because each tree optimization is K times more difficult than binary classification, many implementations use Newton step instead of simple gradient for leaves vectors computation to reduce the number of required steps $T$. The Newton step requires Hessian computation which is quadratic by number of classes and hardly possible when $K$ is big enough. In our approach the optimization at this step is way easier and there is no reason to save the number of gradient steps. Because of this we can skip Newton step to be able to work with large number of classes.

\subsection{Main algorithm}

We will look for decision in the class of such functions:
\begin{equation}
H_c(x) = \sum_{t=1}^T b_{ct} h_t(x)
\end{equation}
It consists of $T$ real-valued components $h_t(x)$ common for all classes. Each component is weighted for each class $c$ differently by $b_{t}$ vector values. The exact probability of the class $c$ is then calculated in form of multinomial logistic regression (\ref{eq:MLRP}). This form allows us to work without conditions on $b_t$ which has $K-1$ components. One could think of $B = \{b_{ct}\}$ as a ``coding matrix'' components of $\{h_t\}_{t=1}^T$ ensemble.

As we want to optimize (\ref{eq:MLR_boosting_step}) for all classes simultaneously then the boosting optimization step (\ref{eq:MLR_boosting_step}) can be split into two steps:
\begin{equation}\begin{array}{l}
(r_{t+1}, b_{t+1}) = \arg \min\limits_{r,b} \|rb^T  - \nabla L_t\|_F \\
h_{t+1} = \arg \min\limits_{h} \|h(X) - r_{t+1}\|_2^2
\end{array}\end{equation}
According to the Eckart-Young-Mirsky theorem \citep{Eckart1936}, solving first step means finding the left and the right singular vectors of $\nabla L_t$ associated with the largest singular value of $\nabla L_t$. The second step is just a regression problem, that is well known and hardly optimized in GBDT libraries. The main steps of the resulting algorithm are summarized in Algorithm.~\ref{alg:fmcb}.

\begin{algorithm}[tb]
\caption{FMCBoosting}
\label{alg:fmcb}
\begin{algorithmic}
\STATE {\bfseries Input:} step $\alpha$, iterations count $T$.
\STATE $H^{(0)}(x):= \mathbb{O} \in \mathbb{F}^{K-1}$ \COMMENT{initial zero model}
\STATE $\overline{x}^{(0)} := \mathbb{O} \in \mathbb{R}^{N \times (K-1)}$ \COMMENT{initial cursor}
 
\FOR{$t=0$ {\bfseries to} $T$}
    \STATE Evaluate the gradient's $\nabla L_{t+1} = \nabla L(\overline{x}^{t})$ using (\ref{eq:grad_matrix}).
    \STATE Factorize the gradient's matrix:
            $$(r_{t+1}, b_{t+1}) = \arg\min_{r,b} \|rb^T - \nabla L_t\|_F^2$$
    \STATE Train weak model $h_{t+1}(x)$ using $\{X,r_{t+1}\}$ as a training set and MSE as a target function:
             $$h_{t+1}(x) = \arg\min_{h} \|h(X) - r_{t+1}\|_2^2$$      
    \STATE Update model: $H^{(t + 1)}(x) = H^{t}(x) + \alpha b_{t+1} h_{t + 1}(x)$.
    \STATE Update cursor: $\overline{x}^{(t+1)} = \overline{x}^{t} + \alpha b_{t+1} h_{t+1}(X)$.
\ENDFOR    
\end{algorithmic}
\end{algorithm}

There are two theoretical issues with the proposed method:
\begin{itemize}
    \item in boosting process neighboring gradients can be aligned with each other and continuous rank-one factorization may become less effective;
    \item we have reduced the complexity of the tree learning procedure, but the SVD cost could take even more time to complete.
\end{itemize}

\subsection{Convergence analysis}
To the best of our knowledge, the multinomial logistic regression boosting analysis is not presented in the literature. On the other hand it seems that general greedy boosting analysis~\citep{boostingconvergence} can be easily adapted for this case. We can use results of this analysis and the only issue with our method is that it can violate the key assumption: approximate minimization. In this assumption authors want the greedy minimization on each step to be closer to the precise optimization on each step. The problem here is that we have introduced additional step (factorization) and it can bring additional error. If the error increases with the number of steps, the boosting can diverge.

To check this we need to evaluate the quality of factorization through optimization steps. We do this with condition number\footnote{Ratio of the greatest singular value to the lowest one}. The lower condition number, the lower the factorization quality. At start of the optimization the condition number drops but with the number of iterations it grows back. Consequently the problem exists only in the first part of optimization, and after a certain point it fades out.

This effect can be explained: at the firsts steps of the boosting the $\nabla L_t$ are aligned with each other, but closer to the optimum, with the reduction of the norm, consequent $\nabla L_t$ become more independent on each other and rank-one factorization becomes effective again.

\subsection{Stochastic ALS}\label{sec:stochasticals}
Rank one matrix factorization is well studied topic. There are many algorithms to deal with this problem, starting from power iteration method to more complex ALS. The ALS complexity is $O(K |\mathbf{X}|)$. In our case, $|\mathbf{X}|$ is the number of observations which we want to grow and this complexity does not fit our needs. It is reasonable to find more effective algorithm for our special case when the number of classes is orders of magnitude less than the number of examples in the dataset $K \ll |\mathbf{X}|$. This algorithm can be used beyond the proposed method so we will describe in general terms of matrix rank one factorization task $\arg \min\limits_{u,v} \|A - u v^T\|$ where $A \in \mathbb{R}^{m\times n}$. Taking into account that elements of $u$ (and their optimization) are independent on each other we can turn our task to stochastic form:

\begin{equation}\begin{array}{l}
\arg \min\limits_{u,v} \|A - u v^T\| = \\\arg \min\limits_{u,v} \sum_{i=1}^m \sum_{j=1}^n (a_{ij} - u_i v_j)^2 = \\
\arg \min\limits_v E_{i \sim U(1, m)}\left(\min\limits_{u_i} \sum_{j=1}^n (a_{ij} - u_i v_j)^2\right)
\end{array}\end{equation}
This task has an infinite number of solutions because we can simultaneously scale $u$ and $v$ by any $\alpha$ and $\frac{1}{\alpha}$ to get the same result. We will look for solutions in set of $\|v\| = 1$. One can see that the third form of the task split it into two independent optimizations of $v$ vector and for $u_i$ component of the $u$ vector. The optimal $u$ can be taken from zero derivative condition of the most internal component as $\hat{u}_i = \frac{\sum_j a_{ij}v_j}{\sum_j v_j^2}$, in condition of the unit length of $v$ it can be rewritten as $\hat{u_i} = \sum_j a_{ij} v_j$. And the $v$ optimization now takes the following form:
\begin{equation}\begin{array}{l}
\arg \min\limits_{v, \|v\| = 1} E_{i \sim U(1, m)}\left(\sum_j (a_{ij} - \hat{u}_i v_j)^2\right)
\end{array}\end{equation}
This task can be easily solved by any stochastic gradient decent method (SGD). We chose the most straightforward way:
\begin{equation}\begin{array}{l}
\hat{v}_j = v_{tj} - 2 w \hat{u}_i (v_{tj} \hat{u}_i - a_{ij}) \\
v_{t+1} = Prj_{\|v\| = 1}(\hat{v})
\end{array}\end{equation}
where $w$ is SGD step. The result $\hat{u}$ is calculated after optimization converges. This algorithm is much faster than the standard ALS on the matrices of the defined shape. When the number of observations is large and only a fixed part of this data $m$ is needed to reach low variance on weak model optimization, the complexity of the SALS algorithm drops to $O(n)$.

\section{Experiments}
\label{experiments}
 In this section, we make two series of experiments: comparing with off-the-shelf libraries and testing the proposed approach on top of the same GBDT implementation. The first part of experiments allows us to validate the statement that the proposed method can achieve the same or even better level of quality in most cases. The second study is dedicated to deeper complexity analysis and the ability of the method to build models that are applicable in industry practice.

\begin{table}[t]
\caption{Statistics for the classification datasets.}
\label{tab:datasets}
% \vskip 0.15in
\begin{center}
\begin{small}
\begin{sc}
\begin{tabular}{lrrr}
\toprule
Data set & Examples & Features & Classes \\
\midrule
segmentation        & 2310      & 19    & 7     \\
abalone             & 4177      & 8     & 29    \\
winequality         & 4898      & 11    & 7     \\
pendigits           & 10992     & 16    & 10    \\
letter              & 20000     & 16    & 26    \\
% mnist               & 70000     & 785   & 10    \\
aloi                & 108000    & 128   & 1000  \\
covertype           & 581012    & 54    & 7     \\
imat2009            & 97290     & 245      & 5 \\
\bottomrule
\end{tabular}
\end{sc}
\end{small}
\end{center}
\vskip -0.1in
\end{table}

\begin{table*}[t]
\caption{Accuracy scores for the Support Vector Machine (SVM), XGBoost, LightGBM, CatBoost, Factorized MultiClass Boosting (FMCB) models on benchmark datasets.}
\label{tab:all_accuracy}
\vspace*{-7pt}
% \vskip 0.15in
\begin{center}
\begin{sc}
\resizebox{\columnwidth}{!}{%
\footnotesize{\begin{tabular}{lccccc}
\toprule
Dataset & SVM  & XGBoost & LightGBM & CatBoost & FMCB  \\
\midrule
segmentation & 0.960 $\pm$ 0.006 & 0.983 $\pm$ 0.005 & 0.982 $\pm$ 0.007 & 0.984 $\pm$ 0.004 & \textit{0.986 $\pm$ 0.001} \\
abalone & 0.268 $\pm$ 0.004 & 0.261 $\pm$ 0.014 & 0.261 $\pm$ 0.009 & \textit{0.276 $\pm$ 0.012} & 0.261 $\pm$ 0.014 \\
winequality & 0.593 $\pm$ 0.019 & 0.676 $\pm$ 0.011 & 0.636 $\pm$ 0.017 & 0.676 $\pm$ 0.005 & \textit{0.684 $\pm$ 0.017} \\
pendigits & 0.993 $\pm$ 0.001 & 0.988 $\pm$ 0.001 & 0.993 $\pm$ 0.001 & \textit{0.994 $\pm$ 0.001} & 0.992 $\pm$ 0.001 \\
letter & 0.957 $\pm$ 0.004 & 0.965 $\pm$ 0.001 & 0.965 $\pm$ 0.001 & 0.966 $\pm$ 0.002 & \textbf{0.969 $\pm$ 0.002} \\
% mnist & 0.984 $\pm$ 0.001 & 0.981 $\pm$ 0.001 & 0.978 $\pm$ 0.001 & 0.981 $\pm$ 0.001 & 0.000 $\pm$ 0.000 \\
aloi & \textbf{0.958 $\pm$ 0.001} & 0.953 $\pm$ 0.001 & 0.952 $\pm$ 0.001 & 0.955 $\pm$ 0.001 & 0.948 $\pm$ 0.001 \\
covertype & ---\footnotemark[3] & 0.960 $\pm$ 0.001 & \textbf{0.964 $\pm$ 0.001} & 0.960 $\pm$ 0.001 & 0.961 $\pm$ 0.001 \\
imat2009 & 0.541 $\pm$ 0.002 & \textbf{0.626 $\pm$ 0.001} & 0.617 $\pm$ 0.005 & 0.610 $\pm$ 0.005 & 0.618 $\pm$ 0.003 \\
\bottomrule
\end{tabular}}%
}
\end{sc}
\end{center}
\vskip -0.1in
\end{table*}

\begin{table*}[t]
\caption{Accuracy scores with the time (hours:minutes) needed to reach this accuracy for XGBoost, LightGBM, CatBoost, Factorized MultiClass Boosting (FMCB) models on ALOI dataset with the different number of classes.}
\label{tab:aloi_accuracy}
\vspace*{-15pt}
\vskip 0.15in
\begin{center}
\begin{small}
\begin{sc}
\resizebox{\columnwidth}{!}{%
\begin{tabular}{lcccc}
\toprule
Dataset & XGBoost & LightGBM & CatBoost & FMCB  \\
\midrule
aloi100 & 0.975 $\pm$ 0.002 (00:08) & 0.979 $\pm$ 0.001 (00:02) & \textbf{0.982 $\pm$ 0.001} (00:31) & \textbf{0.982 $\pm$ 0.002} \underline{(00:13)}\footnotemark[4] \\
aloi250 & 0.973 $\pm$ 0.001 (00:43) & \textit{0.975 $\pm$ 0.001} (00:13) & \textit{0.975 $\pm$ 0.002} (03:36) & 0.973 $\pm$ 0.002 \underline{(00:25)} \\
aloi500 & 0.964 $\pm$ 0.003 (03:54) & 0.965 $\pm$ 0.002 (01:35) & \textit{0.967 $\pm$ 0.001} (21:13) & 0.964 $\pm$ 0.002 \underline{(02:04)} \\
aloi & 0.953 $\pm$ 0.001 (20:08) & 0.952 $\pm$ 0.001 (06:25) & \textbf{0.955 $\pm$ 0.001} (83:20) & 0.948 $\pm$ 0.001 \underline{(03:04)} \\
\bottomrule
\end{tabular}%
}
\end{sc}
\end{small}
\end{center}
\vskip -0.1in
\end{table*}

In the experiments, we use Monte Carlo cross-validation ($5$ splits) to measure the performance of the classifiers on each dataset. $20\%$ of the total instances in each data set is used for testing, another $20\%$ is used for the parameters tuning and with the rest $60\%$ of the instances of the dataset represent final training data. Parameter tuning is done via grid-search method. All experiments were performed on a system with dual Intel Xeon CPUs at 2.60GHz and 32GB RAM, running on Ubuntu 16.04.5

\subsection{Comparison with off-the-shelf libraries}
We consider various implementations of state-of-the-art classification algorithms according to an extensive comparative study~\citep{SOTA}, where the authors compared the accuracy and time efficiency of many classification methods on 71 different datasets. Here we tested the most popular GBDT libraries and SVM. There are many more methods of the multiclass classification known from the literature. That is why we use popular publicly-available classification datasets from the UCI Machine Learning Repository datasets. This allows reader to compare our results with a particular method. In addition, we used a small dataset (IMAT2009) for ranking from Yandex IMAT’2009 competition. To the best of our knowledge, the used baselines outperform the published boosting methods for all presented datasets as we have thoroughly tuned parameters. Tab.~\ref{tab:datasets} shows the characteristics of the used datasets.

We should notice here, that some libraries work much faster on GPU than on CPU, but the performance of GPU version depends on code optimization and it matters more than the computational difficulty of the training process. Because of this, we use CPU versions of all libraries.

Tab.~\ref{tab:all_accuracy} exhibits a comparison of the tested algorithms performance on the benchmark datasets. Each cell contains an average and standard deviation of the accuracy.
% TODO: observations

Amsterdam Library of Objects dataset\footnote[2]{\url{http://aloi.science.uva.nl/}}\footnotetext[3]{We skipped training SVM classifier on Covertype dataset as training process takes too much time}
\footnotetext[4]{One can track FMCB close to linear dependence of training time on classes count} (ALOI) has two properties that we should consider interpreting the results:
\begin{itemize}
    \item it is perfectly balanced (108 examples of each class);
    \item $0.95$+ accuracy for 1000 classes allows us to call this task easy and it needs no sophisticated model to be used as a decision function.
\end{itemize}
The first property gives the advantage to OvR methods, and the second one to simple models. Unfortunately, we are not aware of another non text-based dataset with such a number of classes available publicly and have to use it to study the dependence of accuracy and time on the number of classes.

We analyze how the accuracy and time performance of the gradient boosting family methods depend on the number of classes in a dataset. We randomly choose $100$, $250$ and $500$ classes from the ALOI, which initially consists of $1000$ different classes for this experiment. Tab.~\ref{tab:aloi_accuracy} shows the results of this experiment.

% TODO: SVM
As one can see the accuracy of all methods is similar for each number of classes, but time to train is different. For example, CatBoost achieves the best results on each step due to more sophisticated training mechanism, but it takes too much time to train the model.

Another interesting observation is that the time required to reach the best level of accuracy increases faster than linear which is counter-intuitive for both one-vs-rest and MLR methods where we need to build one weighting function for each class. This effect can be explained in the following way: the total error grows with each classifier and to adjust this growth we need to reduce each individual classifier error by heavier model. This is not the case for FMCB for which the training time grows almost linearly. 

To study how the quality of the model changes when classes are not balanced, we make one more experiment on ALOI dataset. We manually construct an imbalanced version of the dataset by sampling uniformly distributed portion of the original samples for each class. We compare accuracy on original and imbalanced versions of the ALOI dataset for the LightGBM and our method. The drop of model quality for FMCB is less than $5.7\%$\footnote[5]{$3\%$ of the drop comes from 2 times reduction of the dataset.} while for LightGBM it is $8.8\%$. This demonstrates the well-known property of MLR models to be resistant to class imbalance.

\subsection{Performance analysis}
\label{performance}
\begin{table*}%[t]
\centering
\caption{Accuracy scores for the multinomial logistic regression (MLR), one-vs-rest and FMCB with the fixed number of models.}
\label{tab:quality-fixed-models}
\vspace*{-15pt}
\vskip 0.15in
\begin{small}
\begin{sc}
{\renewcommand{\arraystretch}{1.2}% for the vertical padding
\begin{tabular}{lccccc}
\hline
% \abovespace\belowspace
Dataset & \# Models & MLR & OvR  & FMCB & Size (MB) \\
\hline
\hline
\multirow{3}{*}{letters}
& 3120\footnotemark[6] & 0.915 $\pm$ 0.005 & 0.922 $\pm$ 0.006 & \textbf{0.947 $\pm$ 0.002} & 2.142 \\[-0.5ex]
& 6240 & 0.933 $\pm$ 0.004 & 0.943 $\pm$ 0.004 & \textbf{0.958 $\pm$ 0.003} & 4.284 \\[-0.5ex]
& 9100 & 0.940 $\pm$ 0.005 & 0.950 $\pm$ 0.004 & \textbf{0.962 $\pm$ 0.004} & 6.248 \\[-0.5ex]
\hline
\multirow{3}{*}{mnist}
& 1300 & 0.952 $\pm$ 0.002 & 0.957 $\pm$ 0.002 & \textbf{0.963 $\pm$ 0.002} & 0.733\\[-0.5ex]
& 2600 & 0.963 $\pm$ 0.002 & 0.966 $\pm$ 0.002 & \textbf{0.970 $\pm$ 0.002} & 1.467\\[-0.5ex]
& 4000 & 0.967 $\pm$ 0.002 & 0.970 $\pm$ 0.001 & \textbf{0.973 $\pm$ 0.002} & 2.258\\[-0.5ex]
\hline
\multirow{3}{*}{pendigits}
& 1300 & 0.988 $\pm$ 0.003 & 0.990 $\pm$ 0.003 & \textit{0.992 $\pm$ 0.003} & 0.733 \\[-0.5ex]
& 2600 & 0.990 $\pm$ 0.003 & 0.991 $\pm$ 0.002 & \textit{0.993 $\pm$ 0.003} & 1.467 \\[-0.5ex]
& 4000 & 0.991 $\pm$ 0.003 & 0.992 $\pm$ 0.002 & \textit{0.993 $\pm$ 0.003} & 2.258 \\[-0.5ex]
\hline
\end{tabular}
\vskip -0.1in
} % {\renewcommand{\arraystretch}{1.2}% for the vertical padding
\end{sc}
\end{small}
\end{table*}

\begin{table*}[t]
\caption{Performance measurements for tested methods on LETTER collection.}
\label{tab:performance}
\vspace*{-15pt}
\vskip 0.15in
\begin{center}
\begin{small}
\begin{sc}
\begin{tabular}{lrrrr}
\toprule
Method & Boosting steps & Weak models & Training (s) & Decision (ms) \\
\midrule
MLR boosting    & 4900 & 122500 & $ 980 \pm 22$ & $30.1 \pm 0.2$ \\
OVR             & 4610 & 119860 & $ 875 \pm 22$ & $32.1 \pm 0.2$ \\
FMCB       & 10000 & 10000 & $160 \pm 5$   & $2.7 \pm 0.1$ \\
\bottomrule
\end{tabular}
\end{sc}
\end{small}
\end{center}
\vskip -0.1in
\end{table*}

\footnotetext[6]{The number of models should be divisible to the number of classes for OvR and MLR.}

In a production, one often has service level agreement (SLA), formulated in maximal available time to make a decision, and the resulting model must fit this limitation. Because of this open competition, winner solutions (like those from kaggle.com) are rarely (never) used in industry practice. In the previous sections of the experimental study, we did not care about the model complexity and training time, but focus on the quality maximization despite the model weight. In this section, we limit the budget of the model.

Using our own decision trees optimization mechanism, we compare the performance of the three approaches: one-vs-rest, MLR boosting, and FMCB. This way we remove from a consideration the implementation details, that influence the performance a lot~\citep{LightGBM}. Our tree implementation is very close to the one used in CatBoost~\citep{CatBoost} and published at GitHub as Java sources. 

There are two experimental setups:
\begin{itemize}
    \item limit the number of trees in the model and study the quality of the final model;
    \item set up the target model quality and research the time needed to achieve the same quality level through different approaches.
\end{itemize}
From the Tab.~\ref{tab:quality-fixed-models}, it is clear that within a certain budget on the models count, the proposed approach is significantly better than the alternatives. Tab.~\ref{tab:performance} illustrates the performance details of different methods on the LETTER collection when we fix the same level of model quality. As one can see, FMCB model is 10+ times lighter than alternatives providing the same quality. The performance of the FMCB decision function is much better than one for other methods, and this allows to use it in high load setup. The learning process takes six times less time to reach the same quality. These measurements are made on a small number of classes (26), and the difference grows with the number of classes as one can get from previous experiments.

The complexity of the MLR learning is $O(|\mathbf{X}|K|C|T)$ where $T$ is the number of boosting iterations, $C$---possible conditions set of a decision tree. MLR builds $K - 1$ trees on each step and need every bit of available information from a dataset for this. In our method, we build only one tree on each iteration, and its complexity is $O(|\mathbf{X}|(K + |C|)T)$. Further reduction of the complexity term is possible only if we can skip the gradient calculation step that takes $O(|\mathbf{X}|KT)$.

\section{Conclusions and future work}
In this paper, we presented a new approach to solving the multi-classification problem. It is built on top of modeling matrix and inherits its nice properties, e.g., the number of classes is not connected with the number of basic models in the decision function and the method can be used even in case of a large number of classes. The second root of the presented approach is multinomial logistic regression which gives us the ability to work with unbalanced datasets effectively.

The key feature of the method is that training time and model complexity grows linearly with the number of classes. In the experimental section we have shown that this is not the case for the other methods. This property allows us to say that the proposed method applies to a huge number of classes better than other boosting techniques that we have tested.

The presented method is based on the rank-one factorization of the partial derivatives matrix (observations to classes) which is a hard problem regarding computational difficulty. To overcome this difficulty we have transferred ALS algorithm to stochastic form. The resulting algorithm (SALS) has shown its effectiveness on all datasets used in experimental section and eliminated performance issue at this step. Due to promising results, we can recommend using this algorithm in other settings where rank-one factorization needed and the matrix has one dimension dominance over the other.

The resulting decision function is computationally more lightweight than those for existing methods due to its single ensemble nature. The computational difficulty of learning procedure is comparable to a single one vs. rest classification optimization because of SALS algorithm but requires more steps to achieve comparable results. The performance of the proposed method implementation is better than most of the out-of-shelf libraries despite Java-based implementation which is usually slower than the C++ one.

We see the following directions for further improvements:
\begin{itemize}
    \item reducing factorization error;
    \item applying of the presented approach in the multi-label setting;
    \item implementing our method on GPUs.
\end{itemize}
The Java implementation of the algorithm is available at GitHub.

\bibliography{main}
\end{document}